\documentclass[conference]{IEEEtran}
\IEEEoverridecommandlockouts
\usepackage{cite}
\usepackage{amsmath,amssymb,amsfonts}
\usepackage{algorithmic}
\usepackage{graphicx}
\usepackage{textcomp}
\usepackage{xcolor}
\usepackage{footmisc} 
\usepackage{subcaption}
\usepackage[hidelinks]{hyperref}
\usepackage[T1]{fontenc}
\usepackage{url}
\usepackage{fancyhdr}
\usepackage{ragged2e}
\def\BibTeX{{\rm B\kern-.05em{\sc i\kern-.025em b}\kern-.08em
    T\kern-.1667em\lower.7ex\hbox{E}\kern-.125emX}}

\usepackage{xspace}
\newcommand{\Paragraph}[1]{\smallskip\noindent\textbf{#1}\xspace}

\fancypagestyle{firstpage}{
  \fancyhf{} 
  \fancyfoot[C]{%
    \footnotesize
      \justifying
      © 2025 IEEE. Personal use of this material is permitted. Permission from IEEE must be obtained for all other uses, in any current or future media, including reprinting/republishing this material for advertising or promotional purposes, creating new collective works, for resale or redistribution to servers or lists, or reuse of any copyrighted component of this work in other works.
  }
}

\begin{document}

\title{Multi-criteria Rank-based Aggregation for Explainable AI}
\author{
\IEEEauthorblockN{
Sujoy Chatterjee\IEEEauthorrefmark{1}\IEEEauthorrefmark{2},
Everton Romanzini Colombo\IEEEauthorrefmark{2},
Marcos Medeiros Raimundo\IEEEauthorrefmark{2}
}
\IEEEauthorblockA{\IEEEauthorrefmark{1}SR University, Telangana, India\\
sujoychatt24@gmail.com}
\IEEEauthorblockA{\IEEEauthorrefmark{2}Universidade Estadual de Campinas (UNICAMP), São Paulo, Brazil\\
e257234@dac.unicamp.br, mrai@unicamp.br}
}

\maketitle

\thispagestyle{firstpage} 

\begin{abstract}
Explainability is crucial for improving the transparency of black-box machine learning models. With the advancement of explanation methods such as LIME and SHAP, various XAI performance metrics have been developed to evaluate the quality of explanations. However, different explainers can provide contrasting explanations for the same prediction, introducing trade-offs across conflicting quality metrics. Although available aggregation approaches improve robustness, reducing explanations’ variability, very limited research
employed a multi-criteria decision-making approach. To address this gap, this paper introduces a multi-criteria rank-based weighted aggregation method that balances multiple quality metrics simultaneously to produce an ensemble of explanation models. Furthermore, we propose rank-based versions of existing XAI metrics (complexity, faithfulness and stability) to better evaluate ranked feature importance explanations. Extensive experiments on publicly available datasets demonstrate the robustness of the proposed model across these metrics. Comparative analyses of various multi-criteria decision-making and rank aggregation algorithms showed that TOPSIS and WSUM are the best candidates for this use case.
\end{abstract}


\section{Introduction}

The emphasis in the machine learning community has changed to deep learning models \cite{ PPR:PPR379396}. Even with improved results, explanations \cite{Antwarg2021Explaining} of how deep learning models generate their predictions are still required due to the black-box nature of the models mentioned above. Because the General Data Protection Regulation (GDPR) \cite{Goodman_2017} of the European Union mandates that machine learning models must explain their forecasts by the ``right to explanation" act \cite{10.1145/3633598.3633604,NEURIPS2023_89beb2a3,bueff:2022,Ričards:2023}. 

Among explainable methods, Shapley Additive exPlanations (SHAP), Local Interpretable Model-agnostic Explanations (LIME), and Gradient-weighted Class Activation Mapping (Grad-CAM) for neural networks are the most known and commonly trusted \cite{Mitruț:2024}. However, these methods have a post-hoc nature, providing explanations after a prediction generated by a complex black-box model. Despite helping uncover the behavior of the black-box model, they are not linked to the actual model computation. Rudin on 2019 \cite{rudin2019stopexplainingblackbox} emphasizes that post-hoc explanations are frequently faulty and may result in erroneous conclusions on generating forecasts. Also, identical predictions can have conflicting and diverse explanations from many explainable methods \cite{satyapriya:2022,Mitruț:2024}. Depending on the sample and different perturbation techniques, LIME's explanations may present different feature importance scores for similar predictions \cite{Mitruț:2024}. Thus, the interpretability process is hampered by the disparities in approaches, making it challenging for stakeholders to depend on a single explanation. 

To improve explanation quality, some methods aggregate multiple explanations to create a more robust explanation \cite{Rieger2019AggregatingEM, David:2018,Ilse:2019}. However, explanations can be evaluated according to multiple criteria such as fidelity, stability, and complexity \cite{bhatt:2020,David:2018}, and those aggregation methods do not consider the problem's multi-criteria nature. To understand the impact of multi-criteria aggregation, we tested multiple MCDM (multi-criteria decision-making) algorithms (TOPSIS -- Technique for Order of Preference by Similarity to Ideal Solution \cite{Hwang:1981}, and EDAS -- Evaluation Based on Distance from Average Solution \cite{keshavarz:2015} being the most prominent ones) to weigh each explanation's performance. With this information, we employ rank aggregation methods, such as weighted sum (WSUM) \cite{Shengli:2002}, weighted Condorcet (Condorcet) \cite{Montague:2002}, and weighted BordaFuse \cite{Aslam:2001}, to combine multiple component explanations into one single aggregate explanation which integrates multiple explanation methods' contributions, weighed by their performance.

The main contributions of the paper are as follows:

\begin{itemize}
\item We developed a rank-based, multi-criteria aggregation method for local explanation models.\footnote{\url{https://github.com/hiaac-finance/xai\_aggregation}}.
\item We adapted existing XAI (Explainable AI) evaluation metrics for fidelity, stability, and complexity into rank-based versions, so that they work with importance rankings instead of impact (positive and negative values that explain the impact in the prediction) or importance (absolute values that measure how much changing that feature impacts the prediction in either sense) scores.
\item We adapted stability metrics to calculate the impact of in-distribution samples using auto-encoders
\item An experimental analysis that evaluates the impact of MCDM with rank-based aggregation compared to individual explanations. This analysis was conducted in five datasets, comparing three other explanation models against three metrics.
\end{itemize}

\section{Related Works}

Explainable machine learning has seen a surge in techniques to improve transparency and interpretability.  Early work focused on establishing frameworks for understanding model behavior, such as the TAX4CS framework proposed by Szepannaek (2023) \cite{BuckerSGB:22}, which categorizes techniques like partial dependence plots and variable importance as aspects of global explainability.  Other efforts, like Lange et al. (2022) \cite{deLange:2022}, have combined models like LightGBM with SHAP (Shapley Additive exPlanations) to enhance both performance and interpretability by providing insights into variable importance.  More recently, advancements have been made in addressing the Rashomon effect \cite{Mitruț:2024}, which highlights inconsistencies among different explainers or even within a single explainer across instances.  This has led to the development of methods that aggregate explanations from multiple sources (e.g., SHAP, LIME, ANCHOR) and employ techniques like textual-case-based reasoning and cluster-based aggregation to identify meaningful patterns and address discrepancies.

The aggregation of local explanations to gain global insights has also been an active area of research.  Ilse et al. (2019) \cite{Ilse:2019} introduced GALE (global aggregations of local explanations) to bridge the gap between local and global interpretability, examining explanations from models trained on multiclass document classification and binary sentiment analysis tasks using LIME.  Further developments include hierarchical aggregation methods like GLocalX \cite{Setzu:2021} and axiomatic combinations of multiple aggregators using indices like Responsibility Index, Deegan-Packel Index, and Holler-Packel Index \cite{Biradar:2024}. Other approaches, such as AdViCE \cite{Gomez:2020}, leverage aggregated visual counterfactual explanations to understand feature manipulation and its impact on model predictions. 

Despite the wide popularity of various explainable models, there are limitations about the generalized metric to compute the quality of the explanations while aggregating the explanations. Recently, a few approaches addressed this problem \cite{bhatt:2020} and identified several metrics like faithfulness, complexity, and sensitivity \cite{David:2018,Kadir:2023,ancona:2019}. From a philosophical point of view, some research stresses explanations based on the minimal models, and it should not be complex \cite{Batterman:2014}. Similarly, the explanation function should remain unaffected under perturbation, especially when the model output is not changed, and some authors put forward this notion in past research \cite{lipton2003}. Prior research proposed to verify explanation functions from the perspective of feature selection\cite{camburu:2019,Agree:2023}. A recent approach provides a strategy aggregating multiple local explainers with supported theoretical contributions \cite{Alon:2024}. Another method achieves the optimal weights while aggregating multiple explanations, employing the convex combination technique \cite{decker:2024}.

While these methods offer valuable contributions to explainable machine learning, our work distinguishes itself in several key aspects.  Firstly, we adopt a multi-criteria decision-making (MCDM) perspective, which is not considered in the reviewed literature.  Secondly, we utilize an autoencoder-based approach to generate noisy data, differing methodologically from techniques relying solely on random noise. Furthermore, our work is not limited to specific model types like neural networks or tree-based classifiers, as seen in T-Explainer \cite{Evandro:2024}, which faces challenges with categorical data and generalizability to tree-classifiers.  Finally, our approach is evaluated using rank-based performance metrics across diverse datasets, which include credit and healthcare data, demonstrating its broader applicability compared to methods with narrower scopes and evaluation metrics, such as those focusing solely on deep learning models for credit scoring \cite{9386102} or specific applications like diabetic patient surveys \cite{Ahmed:2024}.

\section{Precursory details of Explainable AI: LIME, SHAP, and ANCHOR}

Explainable AI (XAI) techniques help us understand complex machine-learning models. This section includes the introductory details of three popular methods: LIME, SHAP, and ANCHOR.

\subsection*{LIME: Local Interpretable Model-Agnostic Explanations}

LIME uses an interpretable model to approximate any machine learning model locally to explain its prediction \cite{ribeiro2016should}. For a specific input instance $\mathbf{x}$, LIME perturbs the instance by generating a set of samples $\mathbf{x}'$ in the vicinity of $\mathbf{x}$. The original model $f(\cdot)$ is employed to forecast the outputs for the perturbed samples. LIME fits a straightforward, interpretable model $g(\cdot)$ (e.g., linear regression) to approximate the local behaviour of $f(\cdot)$ around $\mathbf{x}$. The aim is to reduce the following loss function:
    \[
    \mathcal{L}(f, g, \pi_{\mathbf{x}}) + \Omega(g),
    \]
where $\mathcal{L}$ is the fidelity loss ensuring $g(\cdot)$ closely approximates $f(\cdot)$ near $\mathbf{x}$, $\pi_{\mathbf{x}}$ is a locality kernel assigning weights to perturbed samples based on their neighbours of $\mathbf{x}$, and $\Omega(g)$ measures the complexity of $g(\cdot)$.

\subsection*{SHAP: Shapley Additive Explanations}

SHAP is a framework derived from cooperative game theory that allocates significance levels to features in model predictions \cite{lundberg2017unified}. SHAP illustrates a prediction $f(\mathbf{x})$ as an aggregate of contributions from distinct features:
    \[
    f(\mathbf{x}) = \phi_0 + \sum_{i=1}^d \phi_i,
    \]
where $\phi_0$ represents the baseline value (average model output), and $\phi_i$ denotes the Shapley value of the $i$-th feature.

The Shapley value $\phi_i$ for a feature $i$ is computed as:
    \[
    \phi_i = \sum_{S \subseteq \{1, \dots, d\} \setminus \{i\}} \frac{|S|!(d - |S| - 1)!}{d!} \left[f(S \cup \{i\}) - f(S)\right],
    \]
let $S$ signify a subset of features, $f(S)$ denotes the model output utilizing solely the features in $S$, and $d$ indicates the total number of features. $\phi_i$ measures the difference between keeping the variable $i$ ($f(S \cup \{i\})$) and removing it ($f(S)$) for all possible combinations of variables $S \subseteq \{1, \dots, d\} \setminus \{i\}$.

\subsection*{ANCHOR: High-Precision Explanations}

ANCHOR generates high-precision, model-agnostic explanations by observing rules that effectively illustrate a model's prediction \cite{ribeiro2018anchors}. An ANCHOR is a set of conditions $A$ that guarantees the same prediction for similar instances that satisfy $A$. Formally, for a model $f(\cdot)$ and a prediction $f(\mathbf{x})$, $A$ is an ANCHOR for $\mathbf{x}$ if:
    \[
    \mathbb{P}(f(\mathbf{x}') = f(\mathbf{x}) \mid \mathbf{x}' \text{ satisfies } A) \geq 1 - \epsilon,
    \]
where $\epsilon$ is a user-specified threshold value, ANCHOR emphasizes precision and guarantees that the conditions in the explanation are faithful to the model's behavior.

Since ANCHOR does not give us a feature importance, we adapt their explanations to provide one. Suppose $n^{\text{range}}_i$ consists of the number of samples within the specified range for feature $i$ (extracted from conditions A), and $n$ denotes the total number of samples in the dataset, $C_i = 1 - \frac{n^{\text{range}}_i}{n}$ is the feature importance given by ANCHOR. $C_i$ capture feature importance from feature $i$ because, for a highly important feature, the number of samples inside A can be reduced to suffice $\mathbb{P}(f(\mathbf{x}') = f(\mathbf{x}) \mid \mathbf{x}' \text{ satisfies } A) \geq 1 - \epsilon$, and for non-important feature we would need $n$ or next to $n$ samples to suffice this property.

\section*{Autoencoder Model Description}

An autoencoder learns an accurate representation of data in an unsupervised way, and is especially used for dimensionality reduction or feature extraction. It consists of an encoder network $\mathbf{z} = f_{\text{encoder}}(\mathbf{x})$ that maps $\mathbf{x} \in \mathbb{R}^d$ into a reduced representation of the data $\mathbf{z} \in \mathbb{R}^q$, where $q < d$. Followed by a decode network $\hat{\mathbf{x}} = f_{\text{decoder}}(\mathbf{z})$ that maps $z$ back to the original feature space $\hat{\mathbf{x}} \in \mathbb{R}^d$. To recover input information, the objective functions consist of $L = \frac{1}{N} \sum_{i=1}^{N} \|\mathbf{x}_i - \hat{\mathbf{x}}_i\|_2^2$ (where $N$ is the number of samples in the dataset). If this objective is sufficiently attained, the latent space representation $\mathbf{z}$ (the encoder's output) keeps the input's information.

\section{Multi-Criteria Analysis}

Achieving conflicting criteria is usually unattainable once we have multiple goals for an explanation. MCDM methods indicate how we can attribute importance weights for explanation alternatives that satisfy a selected trade-off between the criteria. Traditionally, these importance weights are used to rank the other options, but in this paper, we combine the explanations respecting those weights.

\subsection{TOPSIS}

The TOPSIS (Technique for Order of Preference by Similarity to Ideal Solution) model is a widely accepted MCDM method~\cite{Hwang:1981}. It considers that the best alternative should have the shortest distance from the Positive Ideal Solution (PIS) and the most distance from the Negative Ideal Solution (NIS). The steps for implementing TOPSIS are as follows:

\subsection*{Step 1: Building of the Decision Matrix}
We assume there are $m$ alternatives and $n$ criteria. Therefore, the decision matrix $D$ is written by:
\[
D = 
\begin{bmatrix}
x_{11} & x_{12} & \cdots & x_{1n} \\
x_{21} & x_{22} & \cdots & x_{2n} \\
\vdots & \vdots & \ddots & \vdots \\
x_{m1} & x_{m2} & \cdots & x_{mn} \\
\end{bmatrix},
\]
Here, each cell $x_{ij}$ denotes the value of alternative $i$ with respect to criterion $j$. 

\subsection*{Step 2: Normalize the Decision Matrix}
The normalized matrix $R$ is calculated as:
\[
r_{ij} = \frac{x_{ij}}{\sqrt{\sum_{i=1}^{m} x_{ij}^2}}, \quad \forall i = 1, 2, \dots, m \; \text{and} \; j = 1, 2, \dots, n.
\]

\subsection*{Step 3: Compute the Weighted Normalized Decision Matrix}

For each criterion, assign weights $w_j$ so that $\sum_{j=1}^{n} w_j = 1$ appears. Here is the weighted normalized value:
\[
v_{ij} = w_j \cdot r_{ij}, \quad \forall i = 1, 2, \dots, m \; \text{and} \; j = 1, 2, \dots, n.
\]

\subsection*{Step 4: Determine the Positive and Negative Ideal Solutions}
The Positive Ideal Solution (PIS), $A^+$, and the Negative Ideal Solution (NIS), $A^-$, are defined as:
\[
A^+ = \{v_1^+, v_2^+, \dots, v_n^+\}, \quad A^- = \{v_1^-, v_2^-, \dots, v_n^-\},
\]
where $v_j^+ = \max(v_{ij})$ for benefit criteria and $v_j^- = \min(v_{ij})$ for cost criteria.

\subsection*{Step 5: Find the Measures of Separation}
The separation from the PIS and NIS for each alternative is calculated as follows:
\[
S_i^+ = \sqrt{\sum_{j=1}^{n} (v_{ij} - v_j^+)^2}, \quad S_i^- = \sqrt{\sum_{j=1}^{n} (v_{ij} - v_j^-)^2}.
\]

\subsection*{Step 6: Calculate the Relative Closeness to the Ideal Solution}
The relative closeness of each alternative to the PIS is given by (a higher value of $C_i$ indicates a better alternative):
\[
C_i = \frac{S_i^-}{S_i^+ + S_i^-}, \quad 0 \leq C_i \leq 1.
\]

\subsection*{Step 7: Rank the Alternatives}
The $C_i$ values are employed to determine the rank of the alternatives in descending order.

\subsection{EDAS}

Another widely accepted MCDM explored in this research is EDAS (Evaluation Based on Distance from Average Solution)~\cite{keshavarz:2015}, which has a similar alternative to TOPSIS. Still, the average solution is employed as a reference instead of using the positive and negative ideal solutions. The computational steps of the EDAS method are summarized as follows:

\begin{itemize}
    \item Step 1: Construct the decision matrix $X = [x_{ij}]_{m \times n}$, $x_{ij}$ is the performance of alternative $i$ on criterion $j$.
    \item Step 2: Compute the average solution vector $AV = [\bar{x}_j]$, where $\bar{x}_j = \frac{1}{m} \sum_{i=1}^{m} x_{ij}$.
    \item Step 3: Calculate the Positive Distance from Average ($PDA_{ij}$) and Negative Distance from Average ($NDA_{ij}$) matrices. \\
    For benefit criteria ($j \in C_b$): 
    $$
    PDA_{ij} = \max(0, x_{ij} - \bar{x}_j),\\
    NDA_{ij} = \max(0, \bar{x}_j - x_{ij})\\
    $$
    For cost criteria ($j \in C_c$): 
    $$
    PDA_{ij} = \max(0, \bar{x}_j - x_{ij}) ,\\
    NDA_{ij} = \max(0, x_{ij} - \bar{x}_j)\\
    $$
    \item Step 4: Determine the weighted sums for each alternative $i$ using criterion weights $w_j$:
    $$
    SP_i = \sum_{j=1}^{n} w_j PDA_{ij}, \quad SN_i = \sum_{j=1}^{n} w_j NDA_{ij}
    $$
    \item Step 5: Normalize the weighted sums $SP_i$ and $SN_i$:
    $$
    NSP_i = \frac{SP_i}{\max_k(SP_k)}, \quad NSN_i = 1 - \frac{SN_i}{\max_k(SN_k)}
    $$
    \item Step 6: Compute the final Appraisal Scores ($AS_i$):
    $$
    AS_i = \frac{1}{2} (NSP_i + NSN_i)
    $$
    \item Step 7: Rank the alternatives based on descending values of $AS_i$. The alternative with the highest $AS_i$ is ranked highest.
\end{itemize}

\subsection{WSUM}

The core idea of WSUM (Weighted SUM)~\cite{Shengli:2002} is to assign a weight to each input system (or source) and compute an overall score by summing the weighted scores provided by each system. The main calculation involves:
\begin{itemize}
    \item Assigning weights $w_i$ to each criterion $i$.
    \item Obtaining normalized scores $score_{ij}$ for each alternative $j$ and criterion $i$ using this formula: 
    \[
    score = \frac{unnormalized\_score - min\_score}{max\_score - min\_score}
    \]
    \item Calculating the final score $A_j$ for item $j$ using the formula: 
    \[
    A_j = \sum_{i} w_i \cdot score_{ij}
    \]
    \item Ranking items based on their final score $A_j$.
\end{itemize}

\section{Proposed Methodology}

\Paragraph{Data Preprocessing and Model Training:} The used datasets contained a mix of numerical and categorical features. Categorical variables with a relatively small number of categories were transformed using one-hot encoding, while those with many categories were processed with label encoding. Any missing values in the data were handled by removing the corresponding entries. A standard 80:20 train-test split is performed, after which a predictor is trained on the dataset. A Random Forest model was chosen as a good representation of a complex model. However, it is important to emphasize that the proposed method is model-agnostic and can be applied to any black-box predictive model.

\Paragraph{Applying Explanation Models and Evaluating Their Performance:} After fitting the predictor and selecting a data instance to explain, we apply the chosen explainer models intended for aggregation. For this study, we utilized LIME, SHAP, and ANCHOR, although any feature-importance-based explainer—or one that can be converted to such, like Anchor—could be employed. Once the component explanations are generated, they are evaluated based on complexity, fidelity, and sensitivity. A key contribution of this paper is the implementation of three rank-based metrics, each designed to measure one of these criteria. These newly proposed rank-based metrics are applied to assess the performance of each component explanation. In this case, the evaluation results in a 3 (explainers) × 3 (metrics) matrix, which is then used in subsequent steps.

\Paragraph{Applying an MCDM algorithm:} Once the component explanations are evaluated, a Multi-Criteria Decision Making (MCDM) algorithm assigns weights to each component. The primary use case of an MCDM algorithm is to identify the best candidate(s) from a set of options based on multiple criteria, and it does so by using the available data to allocate weights to each option. In this paper's context, however, the goal is not to select the best explanation but to quantify the overall performance of each explainer with a standardized value. Here, the evaluation metrics for the explanations serve as the criteria for the algorithm, and the resulting weights are used as the standardized performance values. Eight MCDM algorithms were considered: EDAS\cite{keshavarz:2015}, TOPSIS \cite{Hwang:1981}, COPRAS \cite{Zavadskas:1994}, PROMETHEE II \cite{Brans:1990,Brans:1985}, ARAS \cite{TurskisZenonas:2010},  COCOSO \cite{Yazdani:2019}, CODAS \cite{Mehdi:2016}, and MABAC \cite{PAMUCAR:2015}.

\Paragraph{Applying Rank Aggregation:} After applying the MCDM algorithm, the resulting weights are passed onto a weighted rank-aggregation algorithm to combine the component explanations, which are themselves rankings of feature importance. This process produces an aggregate feature-importance ranking, which accounts for each component's contribution, weighted by their performance. For this task, three rank-aggregation algorithms were considered: weighted sum \cite{Shengli:2002}, weighted Condorcet \cite{Montague:2002}, and weighted BordaFuse \cite{Aslam:2001}. Among these, only weighted BordaFuse is inherently rank-based. However, all algorithms were adapted to operate exclusively on rank-based values to align ourselves with the goal of maintaining a purely rank-based procedure. Specifically, we chose the squared inverse of the features' ranks as this value.

\begin{figure}[ht!]
\centering
        \includegraphics[width=0.45\textwidth]{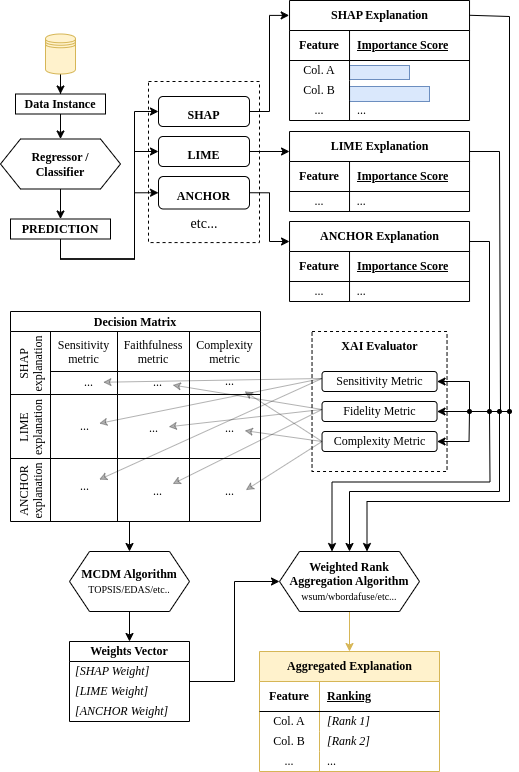}
    \caption{Representation of the workflow in the proposed method.}
    \label{fig:overall_workflow}
\end{figure}

The overall workflow of all of these steps up to here is shown in Fig. \ref{fig:overall_workflow}.

\Paragraph{Validating the Aggregate Explanation:} After obtaining the aggregate explanation, it goes through the same evaluation process as the individual component explanations. By comparing the performance of the aggregate explanation to that of its components, we can assess whether the aggregation procedure yielded satisfactory results. Defining ``satisfactory results" is inherently challenging; however, the primary goal is for the aggregate explanation to avoid achieving the worst value for any of the evaluation metrics.

\subsection{Rank Based Metrics}

A wide range of research has been conducted to identify suitable metrics for evaluating the performance of explanation models \cite{David:2018, Miquel:2025}. In \cite{David:2018}, three relevant evaluation criteria—explicitness, faithfulness, and stability—from diverse perspectives are identified. However, significant challenges remain in the current implementations of metrics that aim to measure these criteria \cite{Miquel:2025}, highlighting the need for new approaches when it comes to XAI evaluation. Aiming for an overall higher robustness, we adopt a rank-based approach in developing these metric implementations.


\paragraph{Normalized Rank-based Complexity (NRC) Metric} Addressing the explicitness criterion, the Normalized Rank-based Complexity (NRC) metric with rank dispersion penalty is defined as follows:

\begin{equation}
\begin{aligned}
&\mu_{NRC}(f, g; \mathbf{x}) = \left( \sum_{i=1}^d \frac{1}{R_i} \right) \cdot \log(d+1) \cdot \left(1 + \alpha \cdot \mathop{\operatorname{std}}(R)\right),  \\
&\text{where } R = r(g(f, x)).
\end{aligned}
\label{Equation:Revised_NRC}
\end{equation}
Here, \( R_i \) is the feature importance rank of the \( i \)-th feature; \( d \) is the total number of features; \( \log(d + 1) \) provides a logarithmic adjustment based on the number of features; \(\text{std}(R)\) is the standard deviation of the feature importance ranks, representing rank dispersion, and \(\alpha\) is a hyperparameter that controls the influence of the rank dispersion penalty. In our experimental analysis,  \(\alpha\) is set to 0.5.

Compared to existing non-rank-based complexity metric implementations, such as the one described in \cite{bhatt:2020}, the NRC metric has the advantage of better capturing the explicitness of explanations that assign nearly identical importance scores to some of the features, while also exhibiting greater resistance to small variations in the explanations.

\paragraph{Rank-based Faithfulness Metric} 
Addressing the fidelity criterion, we chose to adapt the faithfulness metric \cite{bhatt:2020}. Calculating it requires a sample to be disturbed by changing some of its features' values to a baseline value (either zero or the mean value), and then it is observed whether the changes in the prediction caused by this disturbance are correlated to the importance of the changed features given by the explainer model. This metric behaves in a way such that it gives higher values whenever the ``removal" of a highly important feature causes the prediction to change by a relatively higher amount. We present a rank-based implementation of this metric that uses the inverse of a changed feature's rank as a non-parametric measurement of the feature's importance. The formula for this rank-based version is given below. 
\begin{equation}
\begin{split}
    \mu_{F}(f, g; \mathbf{x}) = \mathop{\operatorname{corr}} \limits_{i \in [d]} \left(\frac{1}{r(g(f, \mathbf{x}))_i}, \text{ } |f(\mathbf{x}) - f\left(\mathbf{x}_{[x_i = \hat{x}_i]}\right)| \right) 
\end{split}
\end{equation}
where $d$ is the total number of features, $f(\cdot)$ denotes the predictor function, $g(f,x)$ denotes the local explanation function given a predictor and a data instance, and $r(\cdot)$ signifies the ranking function. Here, the feature $x_i$ is replaced by the baseline value $\hat{x}_i$. 

\paragraph{Rank-based Stability Metric} 
The rank-based stability metric, which addresses the criterion of same name, is calculated by computing the Spearman Correlation between the feature importance scores of a given instance in two contexts: one in which the explainer is fit to the original training data, and another in which the explainer is fit to a noisy variation of the training data. A mathematical model for this metric is shown below.
\begin{equation}
 \mu_{\text{s}}(f, g; \mathbf{x}) = \operatorname{corr}_{\text{spearman}} \bigg( g(f, \mathbf{x} \mid \mathcal{D}), \ g(f, \mathbf{x} \mid \mathcal{D}_{\text{noisy}}) \bigg) 
\end{equation}
Here, $\mathcal{D}$ is the original training dataset, and $\mathcal{D}_{noisy}$ is this same dataset with some noise added to it. $g(f,x|\mathcal{D})$ is the explanation of a given predictor at a specified instance, with the explainer being fit to the provided dataset (which, for the noisy case, is different to the dataset to which the predictor was fit). The rank-based nature of this implementation is ensured by Spearman's correlation, which inherently operates on rank-order calculations.

\textit{Generating noise on tabular data using an autoencoder:} To compute the stability metric, we must obtain a noisy variation of the training dataset. Despite Gaussian noise being a widely accepted method due to its well-defined statistical nature, it presents challenges such as selecting a proper variance value and dealing with categorical features. To address these issues, we employ an autoencoder to assist with generating noise on the data. Given a data sample we wish to perturb, our approach involves identifying its $K$-nearest neighbors w.r.t. the latent space of an autoencoder trained on the dataset. Once these neighbors are determined, we introduce noise by replacing the values of a set number of features with those from the selected neighbors. This method not only ensures that categorical features are handled appropriately, but also provides a more intuitive way to control the noise level: by adjusting the number of neighbors ($K$) and the number of features to be changed rather than tuning variance parameters.

\section{Experimental Design and Results}

Having explained the details regarding the methodology, we want to answer the following research questions:

\begin{itemize}
    \item RQ1. Do the proposed rank-based metrics behave similarly to their traditional counterparts?
    \item RQ2. Does the proposed MCDM rank-based aggregation method create more robust explanations?
    \item RQ3. Do different MCDM algorithms change the aggregation performance?
    \item RQ4. Do different rank aggregation algorithms change the aggregation performance?
\end{itemize}

Before answering those questions, we present experimental details.

\subsection{Dataset Description}
We perform the experiments on three credit scoring datasets, namely, German, Taiwan, and PAKDD2010, and two datasets from other domains: breast cancer~\cite{breast_cancer:17} and student depression~\cite{jaffer:2025}. 
The dataset details and the corresponding features considered for the experimental purposes are listed in Table \ref{table:datasets}. 

\begin{table}[ht!] 
\centering
\caption{Dataset Details}
\begin{tabular}{|c|c|c|c|}
\hline
Datasets & \#samples & \begin{tabular}[c]{@{}c@{}}\#categorical \\ features\end{tabular} & \begin{tabular}[c]{@{}c@{}}\#numerical\\ features\end{tabular} \\ \hline
German \cite{german}   & 1000      & 13                                                                & 7                                                              \\ \hline
Taiwan \cite{taiwan}  & 30,000    & 8                                                                 & 16                                                             \\ \hline
PAKDD2010 \cite{pakdd2010} & 39,988    & 15   & 13                                                             \\ \hline
Breast Cancer \cite{breast_cancer:17}   & 569    & 2                                                               & 28                                                             \\ \hline
Student Depression \cite{jaffer:2025}    & 27,901    & 5                                                                & 13                                                             \\ \hline
\end{tabular}
\label{table:datasets}
\end{table}

\subsection{Experimental Design}
To perform the various experiments, we split the datasets into 80\% and 20\% for training and testing purposes. We trained Random Forest models, which were selected as representations for any complex predictor. The autoencoder that helped compute the stability metric was trained with 500 epochs, using the mean squared error (MSE) as a loss function. For research questions 2 and 3, ten random samples from each dataset were selected, whereas for research question 4, only five random samples were selected, since that lower number was already able to answer the question. Each sample went through the process described in the methodology section. For each experimental run, we identified the individual instance results, and reported only the average results.

\subsection{Validating rank-based metrics}

To answer RQ1, we selected 100 random samples from each credit dataset and applied the three selected explainers—LIME, SHAP, and Anchor—to each sample, resulting in a total of 300 explanations. Our objective was to demonstrate that our newly proposed rank-based metrics effectively capture the same underlying concepts as previously established implementations of XAI evaluation metrics.

For each of the 300 explanations, we computed two sets of evaluation metrics: one using the conventional implementations and another using our rank-based versions. The correlation between each pair of evaluation metric results is presented in Table \ref{Table:Comparison_metrics}.

For complexity, we observe a high Spearman correlation across most datasets, with the exception of PAKDD2010, which shows a low to medium correlation. For faithfulness, all datasets exhibit a strong Spearman correlation. For sensitivity/stability, most datasets display a high negative Spearman correlation, with PAKDD2010 again showing a low to medium correlation. This negative correlation is expected, as the original sensitivity metric interprets lower values as better, whereas the rank-based stability metric follows the opposite interpretation.

\begin{table}[ht!]
\centering
\caption{Correlation between traditional and rank-based metrics using Spearman correlation.}
\begin{tabular}{|c|c|c|c|}
\hline
Datasets & \begin{tabular}[c]{@{}c@{}}Complexity\end{tabular} & \begin{tabular}[c]{@{}c@{}}Faithfulness\end{tabular} & \begin{tabular}[c]{@{}c@{}}Sensitivity\\ Stability\end{tabular} \\ \hline
German   & 0.74                                                            & 0.79            & -0.70                                                \\ \hline
Taiwan   & 0.78                                                            & 0.72        & -0.86                                                    \\ \hline
PAKDD2010    & 0.56                                                            & 0.80         & -0.60                                                   \\ \hline
\end{tabular}
\label{Table:Comparison_metrics}
\end{table}

\ifx
\begin{table}[ht!]
\centering
\caption{Demonstration of similarity between traditional complexity measures Vs. NRC.}
\begin{tabular}{|c|c|c|c|}
\hline
Datasets & \begin{tabular}[c]{@{}c@{}}Spearman\\  Correlation\end{tabular} & \begin{tabular}[c]{@{}c@{}}Pearson \\ Correlation\end{tabular} & \begin{tabular}[c]{@{}c@{}}Kendall \\ Correlation\end{tabular} \\ \hline
German   & 0.74                                                            & 0.67            & 0.51                                                \\ \hline
Taiwan   & 0.78                                                            & 0.80        & 0.58                                                    \\ \hline
PAKDD2010  & 0.56                                                            & 0.30         & 0.41                                                   \\ \hline
\end{tabular}
\label{Table:Comparison_complexity_NRC}
\end{table}

\begin{table}[ht!]
\centering
\caption{Demonstration of similarity between rank-based faithfulness Vs. traditional faithfulness measure}
\begin{tabular}{|c|c|c|c|}
\hline
Datasets & \begin{tabular}[c]{@{}c@{}}Spearman\\  Correlation\end{tabular} & \begin{tabular}[c]{@{}c@{}}Pearson \\ Correlation\end{tabular} & \begin{tabular}[c]{@{}c@{}}Kendall \\ Correlation\end{tabular} \\ \hline
German   & 0.79                                                            & 0.77            & 0.59                                                \\ \hline
Taiwan   & 0.72                                                            & 0.74        & 0.52                                                    \\ \hline
PAKDD2010    & 0.80                                                            & 0.80         & 0.60                                                   \\ \hline
\end{tabular}
\label{Table:Comparison_complexity_faithfulness}
\end{table}
\fi

\subsection{Evaluating the aggregation method} \label{Sec:MCDM}

To answer RQ2, we investigate the performance of the aggregate model when compared to its component explanation models, e.g., LIME, Shap, and ANCHOR. Similarly, to answer RQ3, we investigate the performance of various MCDM models while aggregating the different explanation models. 

Eight MCDM methods (TOPSIS, COPRAS, PROMETHEE II, ARAS, COCOSO, CODAS, EDAS, and MABAC) were evaluated across three credit-scoring datasets. Preliminary experiments showed that TOPSIS and EDAS consistently outperformed the others, thus we only present the results for those. This is mostly due to our unique use case requiring the weights to meaningfully quantify the performance of each component model, rather than simply ranking them (for which MCDM algorithms are usually designed). For each dataset, 10 samples were generated, and the average ranking of each method across various metrics (NRC, stability, and faithfulness) was computed. Lower rankings indicate better performance. 

Tables \ref{Table:instance_avg_TOPSIS_German} and \ref{Table:instance_avg_EDAS_German} present a comparative analysis of different explanation models (LIME, SHAP, ANCHOR, and an aggregate explainer) on the German dataset using TOPSIS and EDAS, respectively. The aggregate explainer performed well across TOPSIS and EDAS, particularly in NRC. SHAP excelled in stability and faithfulness but had a poor NRC. ANCHOR showed strong NRC but weak stability. LIME consistently underperformed in faithfulness. 

\begin{table}[ht!]
\centering
\caption{Results for multiple MCDM methods - German Dataset}
\begin{subtable}[t]{0.5\textwidth}
\centering
\begin{tabular}{|c|c|c|c|}
\hline
Methods             & NRC          & stability    & faithfulness \\ \hline
Aggregate Explainer & \bf{1.6 (2)} & 2.4 (1)      & 2.1 (0)      \\ \hline
ANCHOR              & 1.7 (2)	   & 3.9 (0)      & 2.5 (0)      \\ \hline
LIME                & 3.3 (0)	   & 2.5 (1)      & 3.4 (0)      \\ \hline
SHAP                & 3.4 (0)	   & \bf{1.0 (3)} & \bf{2.0 (0)} \\ \hline
\end{tabular}
\vspace{0.02in}
\caption{TOPSIS - Average rank for 10 instances\protect\footnotemark}
\label{Table:instance_avg_TOPSIS_German}
\end{subtable}
\begin{subtable}[t]{0.5\textwidth}
\centering
\begin{tabular}{|c|c|c|c|}
\hline
Methods             & NRC          & stability    & faithfulness \\ \hline
Aggregate Explainer & \bf{1.4 (2)} & 2.4 (1)      & 2.6 (0)      \\ \hline
ANCHOR              & 1.8 (2)      & 4.0 (0)      & 2.4 (0)      \\ \hline
LIME                & 3.5 (0)      & 2.6 (1)      & 2.7 (0)      \\ \hline
SHAP                & 3.3 (0)      & \bf{1.0 (3)} & \bf{2.3 (0)} \\ \hline
\end{tabular}
\vspace{0.02in}
\caption{EDAS - Average rank for 10 instances\protect\footref{friedman_test}}
\label{Table:instance_avg_EDAS_German}
\end{subtable}
\end{table}
\footnotetext{The number in parentheses shows how many methods are worse than the evaluated one with statistical difference with Finner post hoc (threshold of $p < 0.05$) only after a Friedman test (threshold of $p < 0.05$).\label{friedman_test}}

Tables \ref{Table:instance_avg_TOPSIS_Taiwan}-\ref{Table:instance_avg_EDAS_Taiwan} present the results for the Taiwan dataset (3000 samples after stratified sampling), using both TOPSIS and EDAS. With TOPSIS (Table \ref{Table:instance_avg_TOPSIS_Taiwan}), the aggregate explainer demonstrated strong performance: second-best in NRC, jointly best in faithfulness with ANCHOR, and avoided worst-case performance in stability. Similarly, for EDAS (Table \ref{Table:instance_avg_EDAS_Taiwan}), the aggregate explainer ranked second in NRC, best in faithfulness, and avoided the worst case in stability. While SHAP excelled in stability, the aggregate explainer outperformed the other two metrics. Although ANCHOR had the best NRC, its poor stability makes it less desirable. 

\begin{table}[ht!]
\centering
\caption{Results for multiple MCDM methods - Taiwan Dataset}
\begin{subtable}[t]{0.5\textwidth}
\centering
\begin{tabular}{|c|c|c|c|}
\hline
Methods             & NRC          & stability    & faithfulness \\ \hline
Aggregate Explainer & 1.8 (1)      & 2.9 (0)      & \bf{2.2 (0)} \\ \hline
ANCHOR              & \bf{1.5 (2)} & 4.0 (0)      & \bf{2.2 (0)} \\ \hline
LIME                & 3.8 (0)      & 2.1 (1)      & 3.1 (0)      \\ \hline
SHAP                & 2.9 (0)      & \bf{1.0 (2)} & 2.5 (0)      \\ \hline
\end{tabular}
\vspace{0.02in}
\caption{TOPSIS - Average rank for 10 instances\protect\footref{friedman_test}}
\label{Table:instance_avg_TOPSIS_Taiwan}
\end{subtable}

\begin{subtable}[t]{0.5\textwidth}
\centering
\begin{tabular}{|c|c|c|c|}
\hline
Methods             & NRC          & stability    & faithfulness \\ \hline
Aggregate Explainer & 1.7 (2)      & 2.8 (0)      & \bf{2.2 (0)} \\ \hline
ANCHOR              & \bf{1.4 (2)} & 4.0 (0)      & \bf{2.2 (0)} \\ \hline
LIME                & 3.7 (0)      & 2.2 (1)      & 3.0 (0)      \\ \hline
SHAP                & 3.2 (0)      & \bf{1.0 (2)} & 2.6 (0)      \\ \hline
\end{tabular}
\vspace{0.02in}
\caption{EDAS - Average rank for 10 instances\protect\footref{friedman_test}}
\label{Table:instance_avg_EDAS_Taiwan}
\end{subtable}
\end{table}

Tables \ref{Table:instance_avg_TOPSIS_PAKDD}-\ref{Table:instance_avg_EDAS_PAKDD} show the results for the PAKDD2010 dataset (3000 samples after stratified sampling) using TOPSIS and EDAS. For TOPSIS (Table \ref{Table:instance_avg_TOPSIS_PAKDD}), the aggregate explainer achieved the best NRC and second-best stability, with weaker performance in faithfulness. SHAP performed well in stability and faithfulness but poorly in NRC. With EDAS (Table \ref{Table:instance_avg_EDAS_PAKDD}), the aggregate explainer again had the best NRC and second-best stability. SHAP showed a worse NRC value.

\begin{table}[ht!]
\centering
\caption{Results for multiple MCDM methods - PAKDD2010 Dataset}
\begin{subtable}[t]{0.5\textwidth}
\centering
\begin{tabular}{|c|c|c|c|}
\hline
Methods             & NRC          & stability    & faithfulness \\ \hline
Aggregate Explainer & \bf{1.4 (2)} & 2.0 (1)      & 3.0 (0)      \\ \hline
ANCHOR              & 1.8 (2)      & 3.9 (0)      & 3.0 (0)      \\ \hline
LIME                & 3.4 (0)      & 2.9 (0)      & \bf{1.9 (0)} \\ \hline
SHAP                & 3.4 (0)      & \bf{1.0 (2)} & 2.1 (0)      \\ \hline
\end{tabular}
\vspace{0.02in}
\caption{TOPSIS - Average rank for 10 instances\protect\footref{friedman_test}}
\label{Table:instance_avg_TOPSIS_PAKDD}
\end{subtable}

\begin{subtable}[t]{0.5\textwidth}
\centering
\begin{tabular}{|c|c|c|c|}
\hline
Methods             & NRC          & stability    & faithfulness \\ \hline
Aggregate Explainer & \bf{1.4 (2)} & 2.2 (1)      & 3.1 (0)      \\ \hline
ANCHOR              & 2.0 (1)      & 4.0 (0)      & \bf{2.2 (0)} \\ \hline
LIME                & 3.2 (0)      & 2.8 (0)      & 2.5 (0)      \\ \hline
SHAP                & 3.4 (0)      & \bf{1.0 (2)} & \bf{2.2 (0)} \\ \hline
\end{tabular}
\vspace{0.02in}
\caption{EDAS - Average rank for 10 instances\protect\footref{friedman_test}}
\label{Table:instance_avg_EDAS_PAKDD}
\end{subtable}
\end{table}

Tables \ref{Table:instance_avg_TOPSIS_WDBC}-\ref{Table:instance_avg_EDAS_WDBC} present the results for the breast cancer dataset using TOPSIS and EDAS. With TOPSIS (Table \ref{Table:instance_avg_TOPSIS_WDBC}), the aggregate explainer excelled in NRC and faithfulness while avoiding the worst performance in stability. LIME performed poorly in NRC and faithfulness, while SHAP showed weak NRC despite having the best stability. For EDAS (Table \ref{Table:instance_avg_EDAS_WDBC}), the aggregate explainer again achieved the best NRC and secured second-best results in faithfulness and stability. 

\begin{table}[ht!]
\caption{Results for multiple MCDM methods - WDBC Dataset}
\begin{subtable}[t]{0.5\textwidth}
\centering
\begin{tabular}{|c|c|c|c|}
\hline
Methods             & NRC          & stability    & faithfulness \\ \hline
Aggregate Explainer & \bf{1.1 (2)} & 3.0 (0)      & \bf{1.8 (1)} \\ \hline
ANCHOR              & 2.3 (1)      & 4.0 (0)      & 3.8 (0)      \\ \hline
LIME                & 3.6 (0)      & 2.0 (1)      & 2.5 (1)      \\ \hline
SHAP                & 3.0 (0)      & \bf{1.0 (2)} & 1.9 (1)      \\ \hline
\end{tabular}
\vspace{0.02in}
\caption{TOPSIS - Average rank for 10 instances\protect\footref{friedman_test2}}
\label{Table:instance_avg_TOPSIS_WDBC}
\end{subtable}

\begin{subtable}[t]{0.5\textwidth}
\centering
\begin{tabular}{|c|c|c|c|}
\hline
Methods             & NRC          & stability    & faithfulness \\ \hline
Aggregate Explainer & \bf{1.1 (2)} & 2.5 (1)      & 2.3 (0)      \\ \hline
ANCHOR              & 2.2 (1)      & 4.0 (0)      & 3.4 (0)      \\ \hline
LIME                & 3.7 (0)      & 2.5 (1)      & 2.5 (0)      \\ \hline
SHAP                & 3.0 (0)      & \bf{1.0 (3)} & \bf{1.8 (1)} \\ \hline
\end{tabular}
\vspace{0.02in}
\caption{EDAS - Average rank for 10 instances\protect\footref{friedman_test2}}
\label{Table:instance_avg_EDAS_WDBC}
\end{subtable}
\end{table}

Tables \ref{Table:instance_avg_TOPSIS_Depression_data}-\ref{Table:instance_avg_EDAS_Depression_Data} display the results for the student depression dataset (3486 samples, 10 random samples) using TOPSIS and EDAS. For TOPSIS (Table \ref{Table:instance_avg_TOPSIS_Depression_data}), the aggregate explainer ranked second in NRC and avoided worst-case performance in stability and faithfulness. LIME had the worst NRC, while SHAP showed good stability and faithfulness but poor NRC. ANCHOR had a good NRC but the worst stability. With EDAS (Table \ref{Table:instance_avg_EDAS_Depression_Data}), the aggregate explainer achieved the best faithfulness and second-best NRC, while avoiding the worst stability.

\begin{table}[ht!]
\caption{Results for multiple MCDM methods - Student Depression Dataset}
\begin{subtable}[t]{0.5\textwidth}
\centering
\begin{tabular}{|c|c|c|c|}
\hline
Methods             & NRC          & stability    & faithfulness \\ \hline
Aggregate Explainer & 2.0 (2)      & 3.0 (0)      & 2.6 (0)      \\ \hline
ANCHOR              & \bf{1.0 (2)} & 3.3 (0)      & \bf{2.0 (0)} \\ \hline
LIME                & 3.5 (0)      & 2.6 (0)      & 3.1 (0)      \\ \hline
SHAP                & 3.5 (0)      & \bf{1.1 (3)} & 2.3 (0)      \\ \hline
\end{tabular}
\vspace{0.02in}
\caption{TOPSIS - Average rank for 5 instances\protect\footnotemark}
\label{Table:instance_avg_TOPSIS_Depression_data}
\end{subtable}

\begin{subtable}[t]{0.5\textwidth}
\centering
\begin{tabular}{|c|c|c|c|}
\hline
Methods             & NRC          & stability    & faithfulness \\ \hline
Aggregate Explainer & 1.8 (2)      & 3.3 (0)      & \bf{2.1 (0)} \\ \hline
ANCHOR              & \bf{1.4 (2)} & 3.5 (0)      & 2.6 (0)      \\ \hline
LIME                & 3.3 (0)      & 2.1 (1)      & 3.1 (0)      \\ \hline
SHAP                & 3.5 (0)      & \bf{1.1 (2)} & 2.2 (0)      \\ \hline
\end{tabular}\
\vspace{0.02in}
\caption{EDAS - Average rank for 5 instances\protect\footref{friedman_test2}}
\label{Table:instance_avg_EDAS_Depression_Data}
\end{subtable}
\end{table}

Regarding RQ2, we observe that across multiple datasets (German credit, Taiwan credit, PAKDD2010 credit, breast cancer, and student depression), the aggregate explainer consistently demonstrates strong performance in explaining machine learning models, achieving the best or second-best performance in metrics like NRC and faithfulness, while avoiding worst-case results in stability. Regarding RQ3, our experiments show that TOPSIS and EDAS are somewhat similar.

\footnotetext{The number in parentheses shows how many methods are worse than the evaluated one with statistical difference with Finner post hoc (threshold of $p < 0.05$) only after a Friedman test (threshold of $p < 0.05$).\label{friedman_test2}}

\subsection{Evaluating aggregation algorithms}

To address RQ4, we compared the performance of three rank aggregation algorithms: WSUM, Condorcet, and Borda count. Due to space constraints, only the results for the two most promising techniques, WSUM and Condorcet, are presented (Tables \ref{Table:instance_avg_WSUM_German} - \ref{Table:instance_avg_Condorcet_PAKDD}).

Across all three credit-scoring datasets (German, Taiwan, and PAKDD2010), WSUM consistently outperformed Condorcet, particularly for the aggregate explainer. With WSUM, the aggregate explainer generally achieved the best or second-best performance in NRC, stability, and faithfulness. While SHAP occasionally showed strong performance in individual metrics, its inconsistency made it less desirable. In contrast, Condorcet often yielded less impressive results for the aggregate explainer. These findings suggest that WSUM is a more effective rank aggregation technique for combining multiple explanations, leading to more robust and reliable insights.

\begin{table}[ht!]
\centering
\caption{Results for multiple rank aggregation methods - German Dataset}
\begin{subtable}[t]{0.5\textwidth}
\centering
\begin{tabular}{|c|c|c|c|}
\hline
Methods             & NRC          & stability    & faithfulness \\ \hline
Aggregate Explainer & \bf{1.4 (1)} & 2.0 (0)      & 2.4 (0)      \\ \hline
ANCHOR              & 1.8 (0)      & 3.8 (0)      & 2.4 (0)      \\ \hline
LIME                & 3.6 (0)      & 3.2 (0)      & 3.4 (0)      \\ \hline
SHAP                & 3.2 (0)      & \bf{1.0 (2)} & \bf{1.8 (0)} \\ \hline
\end{tabular}
\vspace{0.02in}
\caption{WSUM - Average rank for 5 instances\protect\footref{friedman_test2}}
\label{Table:instance_avg_WSUM_German}
\end{subtable}

\begin{subtable}[t]{0.5\textwidth}
\centering
\begin{tabular}{|c|c|c|c|}
\hline
Methods             & NRC          & stability    & faithfulness \\ \hline
Aggregate Explainer & 4.0 (0)      & 2.0 (1)      & 2.6 (0)      \\ \hline
ANCHOR              & \bf{1.0 (1)} & 4.0 (0)      & 2.2 (0)      \\ \hline
LIME                & 2.2 (0)      & 3.0 (0)      & 3.2 (0)      \\ \hline
SHAP                & 2.8 (0)      & \bf{1.0 (2)} & \bf{2.0 (0)} \\ \hline
\end{tabular}
\vspace{0.02in}
\caption{Condorcet - Average rank for 5 instances\protect\footref{friedman_test2}}
\label{Table:instance_avg_Condorcet_German}
\end{subtable}
\end{table}

\begin{table}[ht!]
\centering
\caption{Results for multiple rank aggregation methods - Taiwan Dataset}
\begin{subtable}[t]{0.5\textwidth}
\centering
\begin{tabular}{|c|c|c|c|}
\hline
Methods             & NRC          & stability    & faithfulness \\ \hline
Aggregate Explainer & 1.6 (2)      & 2.8 (0)      & 2.8 (0)      \\ \hline
ANCHOR              & \bf{1.4 (2)} & 4.0 (0)      & 2.6 (0)      \\ \hline
LIME                & 3.6 (0)      & 2.2 (0)      & \bf{2.2 (0)} \\ \hline
SHAP                & 3.4 (0)      & \bf{1.0 (1)} & 2.4 (0)      \\ \hline
\end{tabular}
\label{Table:instance_avg_WSUM_Taiwan}
\vspace{0.02in}
\caption{WSUM - Average rank for 5 instances\protect\footref{friedman_test2}}
\end{subtable}

\begin{subtable}[t]{0.5\textwidth}
\centering
\begin{tabular}{|c|c|c|c|}
\hline
Methods             & NRC          & stability    & faithfulness \\ \hline
Aggregate Explainer & 3.8 (0)      & 3.0 (0)      & 3.0 (0)      \\ \hline
ANCHOR              & \bf{1.0 (1)} & 4.0 (0)      & \bf{1.2 (0)} \\ \hline
LIME                & 2.8 (0)      & 2.0 (1)      & 2.6 (0)      \\ \hline
SHAP                & 2.4 (0)      & \bf{1.0 (2)} & 3.2 (0)      \\ \hline
\end{tabular}
\label{Table:instance_avg_Condorcet_Taiwan}
\vspace{0.02in}
\caption{Condorcet - Average rank for 5 instances\protect\footref{friedman_test2}}
\end{subtable}
\end{table}

\begin{table}[ht!]
\centering
\caption{Results for multiple rank aggregation methods - PAKDD2010 Dataset}
\begin{subtable}[t]{0.5\textwidth}
\centering
\begin{tabular}{|c|c|c|c|}
\hline
Methods             & NRC          & stability    & faithfulness \\ \hline
Aggregate Explainer & \bf{1.6 (1)} & 2.25 (0)     & 2.0 (0)      \\ \hline
ANCHOR              & 1.8 (1)      & 3.80 (0)     & 3.6 (0)      \\ \hline
LIME                & 2.8 (0)      & 2.6  (0)     & 2.6 (0)      \\ \hline
SHAP                & 3.8 (0)      & \bf{1.00 (1)}& \bf{1.8 (0)} \\ \hline
\end{tabular}
\vspace{0.02in}
\caption{WSUM - Average rank for 5 instances\protect\footref{friedman_test2}}
\label{Table:instance_avg_WSUM_PAKDD}
\end{subtable}

\begin{subtable}[t]{0.5\textwidth}
\centering
\begin{tabular}{|c|c|c|c|}
\hline
Methods             & NRC          & stability    & faithfulness \\ \hline
Aggregate Explainer & 4.0 (0)      & 2.4 (0)      & 3.0 (0)      \\ \hline
ANCHOR              & \bf{1.4 (1)} & 4.0 (0)      & 2.8 (0)      \\ \hline
LIME                & 2.0 (1)      & 2.6 (0)      & 2.4 (0)      \\ \hline
SHAP                & 2.6 (0)      & \bf{1.0 (1)} & \bf{1.8 (0)} \\ \hline
\end{tabular}
\vspace{0.02in}
\caption{Condorcet - Average rank for 5 instances\protect\footref{friedman_test2}}
\label{Table:instance_avg_Condorcet_PAKDD}
\end{subtable}

\end{table}

We evaluated the proposed method on two healthcare datasets: breast cancer and student depression. For the breast cancer dataset (Tables \ref{Table:instance_avg_WSUM_WDBC} - \ref{Table:instance_avg_Condorcet_WDBC}), the aggregate explainer, using WSUM, achieved the best NRC and faithfulness while avoiding worst-case stability.  With Condorcet, it performed best in faithfulness and avoided the worst stability, but showed a weaker NRC. Similar trends were observed for the student depression dataset (Tables \ref{Table:instance_avg_WSUM_depression} - \ref{Table:instance_avg_Condorcet_depression}). Using WSUM, the aggregate explainer ranked second in NRC and avoided the worst performance in stability and faithfulness. While SHAP had better stability, once more, its poor NRC and faithfulness made it less appealing.

\begin{table}[ht!]
\centering
\caption{Results for multiple rank aggregation methods - WDBC Dataset}
\begin{subtable}[t]{0.5\textwidth}
\centering
\begin{tabular}{|c|c|c|c|}
\hline
Methods             & NRC          & stability    & faithfulness \\ \hline
Aggregate Explainer & \bf{1.0 (2)} & 3.0 (0)      & \bf{1.8 (0)} \\ \hline
ANCHOR              & 2.4 (0)      & 3.8 (0)      & 3.4 (0)      \\ \hline
LIME                & 3.4 (0)      & 2.0 (0)      & 2.6 (0)      \\ \hline
SHAP                & 3.2 (0)      & \bf{1.0 (2)} & 2.2 (0)      \\ \hline
\end{tabular}
\vspace{0.02in}
\caption{WSUM - Average rank for 5 instances\protect\footnotemark}
\label{Table:instance_avg_WSUM_WDBC}
\end{subtable}

\begin{subtable}[t]{0.5\textwidth}
\centering
\begin{tabular}{|c|c|c|c|}
\hline
Methods             & NRC          & stability    & faithfulness \\ \hline
Aggregate Explainer & 4.0 (0)      & 2.8 (0)      & \bf{1.2 (1)} \\ \hline
ANCHOR              & \bf{1.0 (1)} & 4.0 (0)      & 4.0 (0)      \\ \hline
LIME                & 2.4 (0)      & 2.2 (0)      & 2.2 (0)      \\ \hline
SHAP                & 2.6 (0)      & \bf{1.0 (1)} & 2.6 (0)      \\ \hline
\end{tabular}
\vspace{0.02in}
\caption{Condorcet - Average rank for 5 instances\protect\footref{friedman_test3}}
\label{Table:instance_avg_Condorcet_WDBC}
\end{subtable}
\end{table}

\begin{table}[ht!]
\centering
\caption{Results for multiple rank aggregation methods - Student Depression Dataset}
\begin{subtable}[t]{0.5\textwidth}
\centering
\begin{tabular}{|c|c|c|c|}
\hline
Methods             & NRC          & stability    & faithfulness \\ \hline
Aggregate Explainer & 1.8 (0)      & 3.2 (0)      & 2.6 (0)      \\ \hline
ANCHOR              & \bf{1.2 (2)} & 3.4 (0)      & 2.4 (0)      \\ \hline
LIME                & 3.5 (0)      & 2.0 (0)      & \bf{2.2 (0)} \\ \hline
SHAP                & 3.5 (0)      & \bf{1.4 (0)} & 2.8 (0)      \\ \hline
\end{tabular}
\vspace{0.02in}
\caption{WSUM - Average rank for 5 instances\protect\footref{friedman_test3}}
\label{Table:instance_avg_WSUM_depression}
\end{subtable}

\begin{subtable}[t]{0.5\textwidth}
\centering
\begin{tabular}{|c|c|c|c|}
\hline
Methods             & NRC          & stability    & faithfulness \\ \hline
Aggregate Explainer & 4.0 (0)      & 3.0 (0)      & 2.4 (0)      \\ \hline
ANCHOR              & \bf{1.0 (1)} & 3.8 (0)      & \bf{2.2 (0)} \\ \hline
LIME                & 2.4 (0)      & 2.2 (0)      & 2.6 (0)      \\ \hline
SHAP                & 2.6 (0)      & \bf{1.0 (2)} & 2.8 (0)      \\ \hline
\end{tabular}
\vspace{0.02in}
\caption{Condorcet - Average rank for 5 instances\protect\footref{friedman_test3}}
\label{Table:instance_avg_Condorcet_depression}
\end{subtable}
\end{table}

Reinforcing RQ2, the aggregate explainer consistently demonstrated strong and stable performance across all datasets and metrics. Individual explainers, such as SHAP, LIME, and ANCHOR, often exhibited inconsistencies, excelling in some metrics while underperforming in others.  Regarding RQ4, WSUM consistently proved to be a more effective method of rank aggregation than Condorcet. These findings strongly suggest that the aggregate explainer, particularly when paired with WSUM, offers a robust and reliable approach to generating explanations across various domains.

\footnotetext{The number in parentheses shows how many methods are worse than the evaluated one with statistical difference with Finner post hoc (threshold of $p < 0.05$) only after a Friedman test (threshold of $p < 0.05$).\label{friedman_test3}}

\section{Limitations}

In this research, we verified the following limitations. \textbf{Computational Cost:} The computational cost of the proposal is equivalent to the execution of SHAP multiple times (other methods are usually cheaper), increasing the explanation cost to improve explanation robustness; \textbf{NRC Parameter:} $\alpha = 0.5$ was selected to make a good trade-off between the two factors, i.e., lower-ranked features' importance and rank dispersion. However, it was selected based on preliminary experiments to maintain a nominal penalty for high standard deviation; \textbf{Criteria Parameters:} we assumed there is insufficient information or consensus to assign differential weights, thus defining equal weights, which showed an increase in robustness, but further study is needed to understand the impact of differential weights, particularly to improve performance on stability while keeping a good performance in the other metrics; \textbf{Lack of visual explanations:} since the output of the method consists of ranks, the visualization of single explanations does not add information compared with a simple list.

\section{Conclusion}
This work studied the use of multicriteria decision-making methods, rank-based approaches to performance metrics, and rank aggregation algorithms in the context of explainable AI. We proposed a method to aggregate multiple explanation models, aiming to enhance the overall robustness of explanations. To maintain purely rank-based procedures, we developed rank-based versions of existing XAI metrics for complexity, stability, and faithfulness. Once the explanations from the component models were evaluated against these metrics, a MCDM algorithm was used to quantify the performance of each component using scalar weights, which were later combined by a rank aggregation algorithm to form a single explanation. Experiments comparing MCDM and rank aggregation algorithms revealed TOPSIS and WSUM to be the best candidates for this use-case. A comprehensive experimental analysis across five datasets demonstrated the technique's effectiveness in enhancing explanation robustness.

\section*{Acknowledgements}

This project was supported by the Brazilian Ministry of Science, Technology, and Innovation, with resources from Law nº 8,248, of October 23, 1991, within the scope of PPI-SOFTEX, coordinated by Softex and published Arquitetura Cognitiva (Phase 3), DOU 01245.003479/2024-10.

\bibliographystyle{IEEEtran}
\bibliography{references}

\end{document}